\documentclass[11pt,a4paper]{article}
\usepackage[hyperref]{naaclhlt2019}
\usepackage{times}
\usepackage{latexsym}
\usepackage{epsfig}
\usepackage{graphicx}
\usepackage{amsmath}
\usepackage{amssymb}

\usepackage[belowskip=0pt,aboveskip=0pt,font=small]{caption}
\usepackage[belowskip=0pt,aboveskip=0pt,font=scriptsize]{subcaption}
\usepackage{url}

\graphicspath{ {figures/} }
\usepackage{stmaryrd}

\aclfinalcopy 
\def\aclpaperid{1636} 

\title{
Learning to Navigate Unseen Environments: \\Back Translation with Environmental Dropout
}

\author{Hao Tan \;\;\;\;\;\;\; Licheng Yu \;\;\;\;\;\;\; Mohit Bansal \\
  UNC Chapel Hill \\
  {\tt \{haotan, licheng, mbansal\}@cs.unc.edu} \\
 }

\date{}

\begin{document}
\maketitle

\begin{abstract}
A grand goal in AI is to build a robot that can accurately navigate based on natural language instructions, which requires the agent to perceive the scene, understand and ground language, and act in the real-world environment. One key challenge here is to learn to navigate in new environments that are unseen during training. Most of the existing approaches perform dramatically worse in unseen environments as compared to seen ones.
In this paper, we present a generalizable navigational agent.
Our agent is trained in two stages.
The first stage is training via mixed imitation and reinforcement learning, combining the benefits from both off-policy and on-policy optimization.
The second stage is fine-tuning via newly-introduced `unseen' triplets (environment, path, instruction).
To generate these unseen triplets, we propose a simple but effective `environmental dropout' method to mimic unseen environments, which overcomes the problem of limited seen environment variability.
Next, we apply semi-supervised learning (via back-translation) on these dropped-out environments to generate new paths and instructions.
Empirically, we show that our agent is substantially better at generalizability when fine-tuned with these triplets, outperforming the state-of-art approaches by a large margin on the private unseen test set of the Room-to-Room task, and achieving the top rank on the leaderboard.\footnote{Our code, data, and models publicly available at: \url{https://github.com/airsplay/R2R-EnvDrop}}
\end{abstract}

\section{Introduction}
\begin{figure}[t]
\centering
\includegraphics[width=0.50\textwidth]{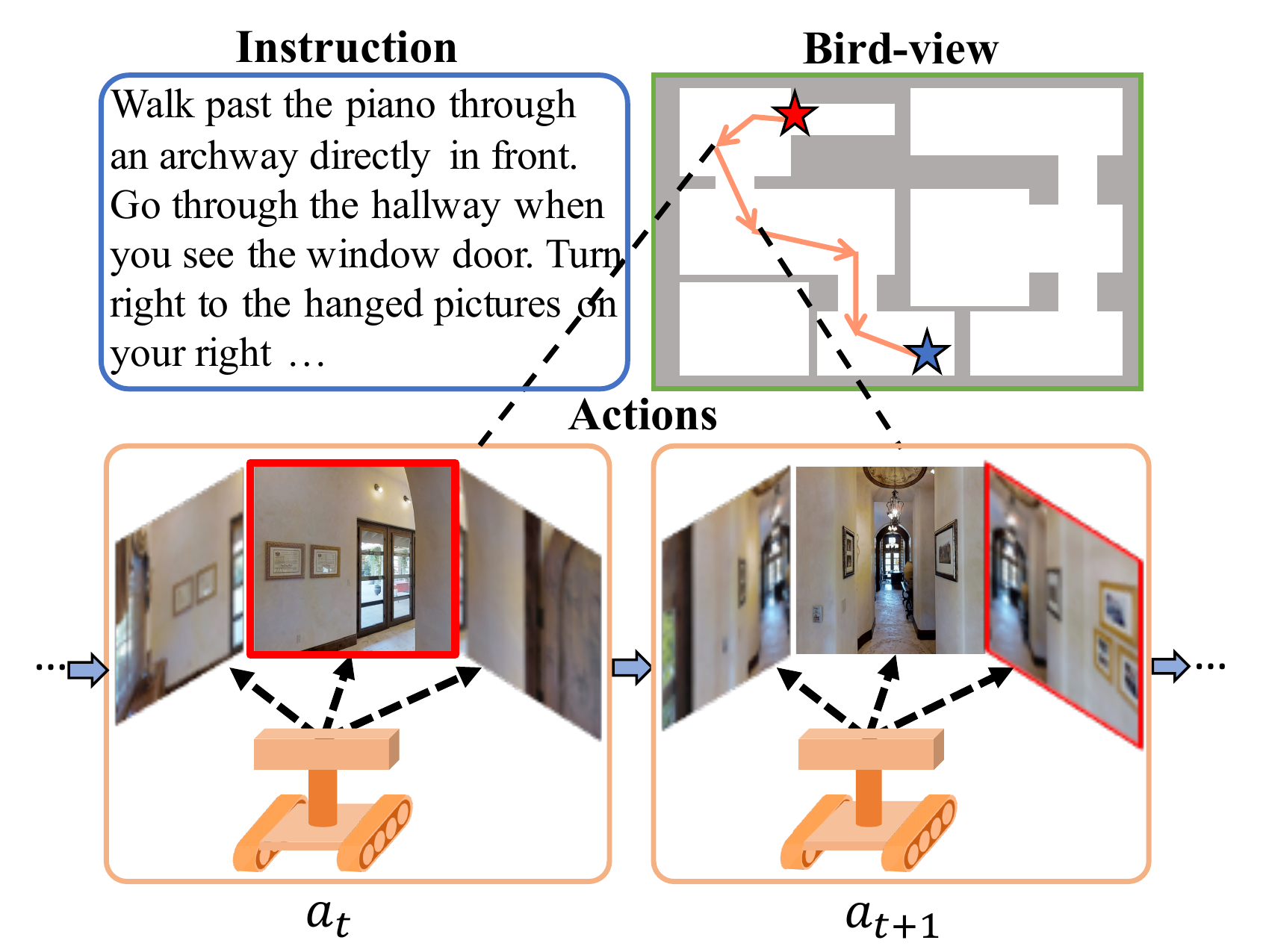}
\vspace{-0.2cm}
\caption{Room-to-Room Task. The agent is given an instruction, then starts its navigation from some staring viewpoint inside the given environment. At time $t$, the agent selects one view (highlighted in red) from a set of its surrounding panoramic views to step into, as an action $a_t$. 
}\label{fig:intro}
\vspace{-0.2cm}
\label{fig:pg}
\end{figure}

One of the important goals in AI is to develop a robot/agent that can understand instructions from humans and perform actions in complex environments. 
In order to do so, such a robot is required to perceive the surrounding scene, understand our spoken language, and act in a real-world house. 
Recent years have witnessed various types of embodied action based NLP tasks being proposed~\cite{correa2010multimodal,walters2007robotic,hayashi2007humanoid, zhu2017target,das2018embodied,mattersim}.

In this paper, we address the task of instruction-guided navigation, where the agent seeks a route from a start viewpoint to an end viewpoint based on a given natural language instruction in a given environment, as shown in Fig.~\ref{fig:intro}.
The navigation simulator we use is the recent Room-to-Room (R2R) simulator~\cite{mattersim}, which uses real images from the Matterport3D~\cite{Matterport3D} indoor home environments and collects complex navigable human-spoken instructions inside the environments, hence connecting problems in vision, language, and robotics.
The instruction in Fig.~\ref{fig:intro} is \emph{``Walk past the piano through an archway directly in front. Go through the hallway when you see the window door. Turn right to the hanged pictures..."}.
At each position (viewpoint), the agent perceives panoramic views (a set of surrounding images) and selects one of them to step into.
In this challenging task, the agent is required to understand each piece of the instruction and localize key views (\emph{``piano"}, \emph{``hallway"}, \emph{``door"}, etc.) for making actions at each time step.
Another crucial challenge is to generalize the agent's navigation understanding capability to unseen test room environments, considering that the R2R task has substantially different unseen (test) rooms as compared to seen (trained) ones.
Such generalization ability is important for developing a practical navigational robot that can operate in the wild.

Recent works~\cite{fried2018speaker, wang2018reinforced,wang2018look,anonymous2019self-aware} have shown promising progress on this R2R task, based on speaker-follower, reinforcement learning, imitation learning, cross-modal, and look-ahead models.
However, the primary issue in this task is that most models perform substantially worse in unseen environments than in seen ones, due to the lack of generalizability.
Hence, in our paper, we focus on improving the agent's generalizability in unseen environments.
For this, we propose a two-stage training approach.
The first stage is training the agent via mixed imitation learning (IL) and reinforcement learning (RL) which combines off-policy and on-policy optimization; this significantly outperforms using IL or RL alone.

The second, more important stage is semi-supervised learning with generalization-focused `environmental dropout'. Here, the model is fine-tuned using additional training data generated via back-translation. This is usually done based on a neural speaker model~\cite{fried2018speaker} that synthesizes new instructions for additional routes in the \emph{existing} environments.
However, we found that the bottleneck for this semi-supervised learning method is the limited variability of given (seen) environments.
Therefore, to overcome this, we propose to generate novel and diverse environments via a simple but effective `environmental dropout' method based on view- and viewpoint-consistent masking of the visual features.
Next, the new navigational routes are collected from these new environments, and lastly the new instructions are generated by a neural speaker on these routes, and these triplets are employed to fine-tune the model training.

Overall, our fine-tuned model based on back-translation with environmental dropout substantially outperforms the previous state-of-the-art models, and achieves the most recent rank-1 on the Vision and Language Navigation (VLN) R2R challenge leaderboard's private test data, outperforming all other entries in success rate under all evaluation setups (single run, beam search, and pre-exploration).\footnote{\url{https://evalai.cloudcv.org/web/challenges/challenge-page/97/overview}} We also present detailed ablation and analysis studies to explain the effectiveness of our generalization method.

\section{Related Work}

\paragraph{Embodied Vision-and-Language}
Recent years are witnessing a resurgence of active vision.
For example, \citet{Levine2016} used an end-to-end learned model to predict robotic actions from raw pixel data, \citet{gupta2017cognitive} learned to navigate via mapping and planning, \citet{sadeghi2017cadrl} trained an agent to fly in simulation and show its performance in the real world, and \citet{gandhi2017} trained a self-supervised agent to fly from examples of drones crashing.
Meanwhile, in the intersection of active perception and language understanding, several tasks have been proposed, including instruction-based navigation~\cite{chaplot2017gated,mattersim}, target-driven navigation~\cite{zhu2017target,gupta2017cognitive}, embodied question answering~\cite{das2018embodied}, interactive question answering~\cite{gordon2018iqa}, and task planning~\cite{zhu2017visual}.
While these tasks are driven by different goals, they all require agents that can perceive their surroundings, understand the goal (either presented visually or in language instructions), and act in a virtual environment.

\paragraph{Instruction-based Navigation}
For instruction-based navigation task, an agent is required to navigate from start viewpoint to end viewpoint according to some given instruction in an environment.
This task has been studied by many works~\cite{tellex2011understanding,chen2011learning,artzi2013weakly,andreas2015alignment,mei2016listen,misra2017mapping} in recent years.
Among them,~\cite{mattersim} differs from the others as it introduced a photo-realistic dataset -- Room-to-Room (R2R), where all images are real ones taken by Matterport3D~\cite{Matterport3D} and the instructions are also natural.
In R2R environments, the agent's ability to perceive real-world images and understanding natural language becomes even more crucial.
To solve this challenging task, a lot of works
\cite{fried2018speaker, wang2018look,wang2018reinforced, anonymous2019self-aware} 
have been proposed and shown some potential. 
The most relevant work to us is \newcite{fried2018speaker}, which proposed to use a speaker to synthesize new instructions and implement pragmatic reasoning.
However, we observe there is some performance gap between seen and unseen environments.
In this paper, we focus on improving the agent's generalizability in unseen environment.

\paragraph{Back-translation}
Back translation~\cite{sennrich2016improving}, a popular semi-supervised learning method, has been well studied in neural machine translation~\cite{hoang2018iterative, wang2018switchout, edunov2018understanding, style_transfer_acl18}.
Given paired data of source and target sentences, the model first learns two translators -- a forward translator from source to target and a backward translator from target to source.
Next, it generates more source sentences using the back translator on an external target-language corpus.
The generated pairs are then incorporated into the training data for fine-tuning the forward translator, which proves to improve the translation performance.
Recently, this approach (also known as data augmentation) was applied to the task of instruction-based navigation~\cite{fried2018speaker}, where the source and target sentences are replaced with instructions and routes.

\begin{figure*}[t]
\centering
\includegraphics[width=0.98\textwidth]{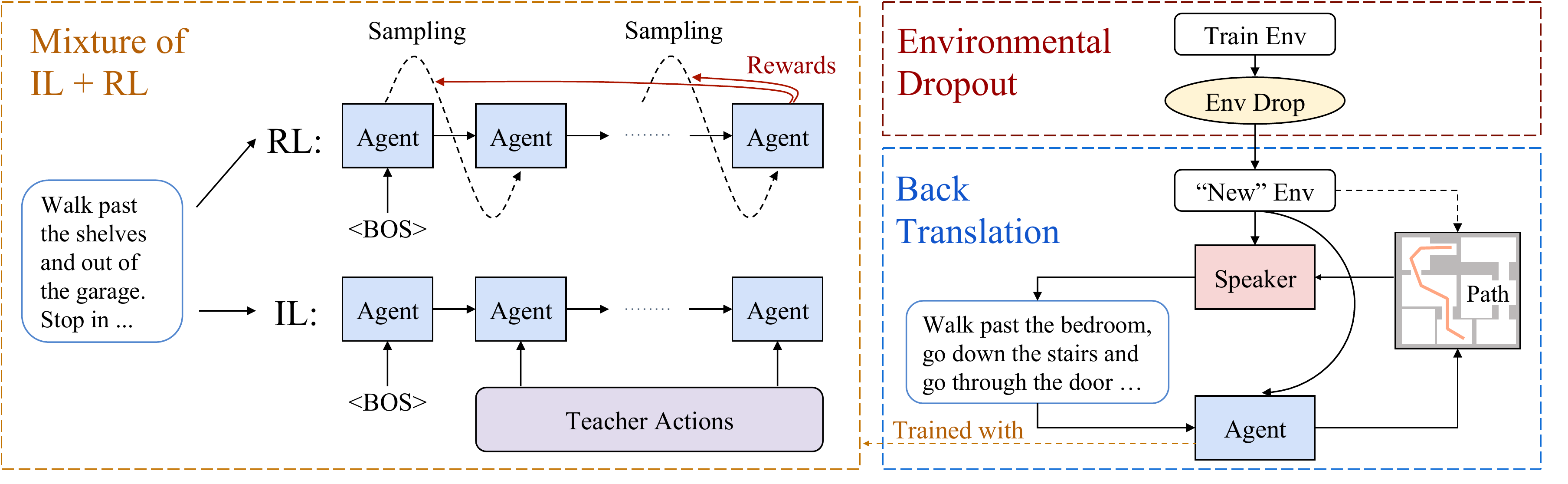}
\vspace{4pt}
\caption{Left: IL+RL supervised learning (stage 1). Right: Semi-supervised learning with back translation and environmental dropout (stage 2). 
}
\vspace{-10pt}
\label{fig:ssl}
\end{figure*}
\section{Method}
\label{sec:method}

\subsection{Problem Setup}
\label{sec:problem_setup}
Navigation in the Room-to-Room task~\cite{mattersim} requires an agent to find a route $\mathbf R$ (a sequence of viewpoints) from the start viewpoint $\mathbf{S}$ to the target viewpoint $\mathbf{T}$ according to the given instruction $\mathbf{I}$.
The agent is put in a photo-realistic environment $\mathbf{E}$. 
At each time step $t$, the agent's observation consists of a panoramic view and navigable viewpoints.
The panoramic view $o_t$ is discretized into $36$ single views $\{o_{t,i}\}_{i=1}^{36}$.
Each single view $o_{t,i}$ is an RGB image $v_{t,i}$ accompanied with its orientation $(\theta_{t,i}, \phi_{t,i})$, where $\theta_{t,i}$ and $\phi_{t,i}$ are the angles of heading and elevation, respectively.
The navigable viewpoints $\{l_{t,k}\}_{k=1}^{N_t}$ are the $N_t$ reachable and visible locations from the current viewpoint.
Each navigable viewpoint $l_{t,k}$ is represented by the orientation $(\hat\theta_{t,k}, \hat\phi_{t,k})$ from current viewpoint to the next viewpoints.
The agent needs to select the moving action $a_t$ from the list of navigable viewpoints $\{l_{t,k}\}$ according to the given instruction $\mathbf{I}$, history/current panoramic views $\{o_{\tau}\}_{\tau=1}^{t}$, and history actions $\{a_{\tau}\}_{\tau=1}^{t-1}$. 
Following \newcite{fried2018speaker}, we concatenate the ResNet~\cite{he2016deep} feature of the RGB image and the orientation as the view feature $f_{t,i}$:
\begin{align}
\vspace{-10pt}
f_{t,i} = [\,&\mathrm{ResNet}(v_{t,i}); \nonumber \\ 
&\left(\cos \theta_{t,i}, \sin \theta_{t,i}, \cos \phi_{t,i}, \sin \phi_{t,i} \right)\, ]
\vspace{-10pt}
\end{align}
The navigable viewpoint feature $g_{t,k}$ is extracted in the same way.

\subsection{Base Agent Model}
\label{sec:model} 
For our base instruction-to-navigation translation agent, we implement an encoder-decoder model similar to \newcite{fried2018speaker}. 
The encoder is a bidirectional LSTM-RNN with an embedding layer:
\vspace{-5pt}
\begin{align}
\hat{w}_j &= \mbox{embedding}(w_j) \\ 
u_1, u_2, \cdots, u_\textsc{L} &= \mathrm{Bi\mbox{-}LSTM} (\hat{w}_1, \cdots, \hat{w}_\textsc{L})
\vspace{-10pt}
\end{align} 
where $u_j$ is the $j$-th word in the instruction with a length of $L$.
The decoder of the agent is an attentive LSTM-RNN. 
At each decoding step $t$, the agent first attends to the view features $\{f_{t,i}\}$ computing the attentive visual feature $\tilde{f}_t$:
\vspace{-5pt}
\begin{align}
\alpha_{t,i} &= \mathrm{softmax}_i(f_{t,i}^\intercal W_\textsc{f} \, \tilde{h}_{t-1}) \\
\tilde f_t &= \sum\nolimits_i \alpha_{t,i} f_{t,i}
\vspace{-5pt}
\end{align}
The input of the decoder is the concatenation of the attentive visual feature $\tilde{f}_t$ and the embedding of the previous action $\tilde{a}_{t-1}$. 
The hidden output $h_t$ of the LSTM is combined with the attentive instruction feature $\tilde u_t$ to form the instruction-aware hidden output $\tilde h_t$. 
The probability of moving to the $k$-th navigable viewpoint $p_t(a_{t,k})$ is calculated as softmax of the alignment between the navigable viewpoint feature $g_{t,k}$ and the instruction-aware hidden output $\tilde h_t$.
\begin{align}
h_t & = \mathrm{LSTM}\left([ \tilde f_t;\tilde a_{t-1}], \tilde h_{t-1}\right) \\
\beta_{t,j} &= \mathrm{softmax}_j \left(u_j ^\intercal \, W_\textsc{u}\, h_t\right) \\
\tilde u_t & = \sum\nolimits_j \beta_{t,j}\, u_j \\
\tilde h_t &= \tanh \left(W \left[\tilde u_t ; \, h_t \right]\right) \\ 
p_t(a_{t,k}) & = \mathrm{softmax}_k \left(g_{t, k}^\intercal \,W_\textsc{G}\, \tilde h_t\right) 
\end{align}
Different from \newcite{fried2018speaker}, we take the instruction-aware hidden vector $\tilde h_{t-1}$ as the hidden input of the decoder instead of $h_{t-1}$. 
Thus, the information about which parts of the instruction have been attended to is accessible to the agent. 
\subsection{Supervised Learning: Mixture of Imitation+Reinforcement Learning}
\label{sec:rl}
We discuss our IL+RL supervised learning method in this section.\footnote{As opposed to semi-supervised methods in Sec.~\ref{sec:ssl}, in this section we view both imitation learning and reinforcement learning as supervised learning.}

\paragraph{Imitation Learning (IL)}
In IL, an agent learns to imitate the behavior of a teacher. 
The teacher demonstrates a teacher action $a^*_t$ at each time step $t$.
In the task of navigation, a teacher action $a^*_t$ selects the next navigable viewpoint which is on the shortest route from the current viewpoint to the target $\mathbf{T}$.
The off-policy\footnote{According to \newcite{poole2010artificial}, an off-policy learner learns the agent policy independently of the agent's navigational actions. An on-policy learner learns the policy from the agent's behavior including the exploration steps.} agent learns from this weak supervision by minimizing the negative log probability of the teacher's action $a^*_t$. 
The loss of IL is as follows:
\begin{equation}
\mathcal{L}^\textsc{IL} = \sum_t \mathcal{L}^\textsc{IL}_t =  \sum_t \mbox{-} \log p_t(a^*_t) 
\end{equation}
For exploration, we follow the IL method of Behavioral Cloning~\cite{bojarski2016end}, where the agent moves to the viewpoint following the teacher's action $a^*_t$ at time step $t$.

\paragraph{Reinforcement Learning (RL)}
Although the route induced by the teacher's actions in IL is the shortest, this selected route is not guaranteed to satisfy the instruction. 
Thus, the agent using IL is biased towards the teacher's actions instead of finding the correct route indicated by the instruction. 
To overcome these misleading actions, the on-policy reinforcement learning method Advantage Actor-Critic \cite{mnih2016asynchronous} is applied,
where the agent takes a sampled action from the distribution $\{p_t(a_{t,k})\}$ and learns from rewards. 
If the agent stops within $3m$ around the target viewpoint $\mathbf T$, a positive reward $+3$ is assigned at the final step. 
Otherwise, a negative reward $-3$ is assigned.
We also apply reward shaping \cite{wu2018building}: the direct reward at each non-stop step $t$ is the change of the distance to the target viewpoint.

\paragraph{IL+RL Mixture}
To take the advantage of both off-policy and on-policy learners, we use a method to mix IL and RL. 
The IL and RL agents share weights, take actions separately, and navigate two independent routes (see Fig.~\ref{fig:ssl}).
The mixed loss is the weighted sum of $\mathcal{L}^\textsc{IL}$ and $\mathcal{L}^\textsc{RL}$:
\begin{equation}
\mathcal{L}^\textsc{MIX} = \mathcal{L}^\textsc{RL} + \lambda_\textsc{IL} \mathcal{L}^\textsc{IL}
\end{equation}
IL can be viewed as a language model on action sequences, which regularizes the RL training.\footnote{This approach is similar to the method ML+RL in \newcite{paulus2017deep} for summarization. 
Recently,~\citet{wang2018look} combines purely supervised learning and RL training however, they use a different algorithm named MIXER~\cite{ranzato2015sequence}, which computes cross entropy (XE) losses for the first $k$ actions and RL losses for the remaining.}

\begin{figure*}[t]
    \begin{subfigure}[t]{0.48\textwidth}
        \centering
        \includegraphics[width=1.0\textwidth]{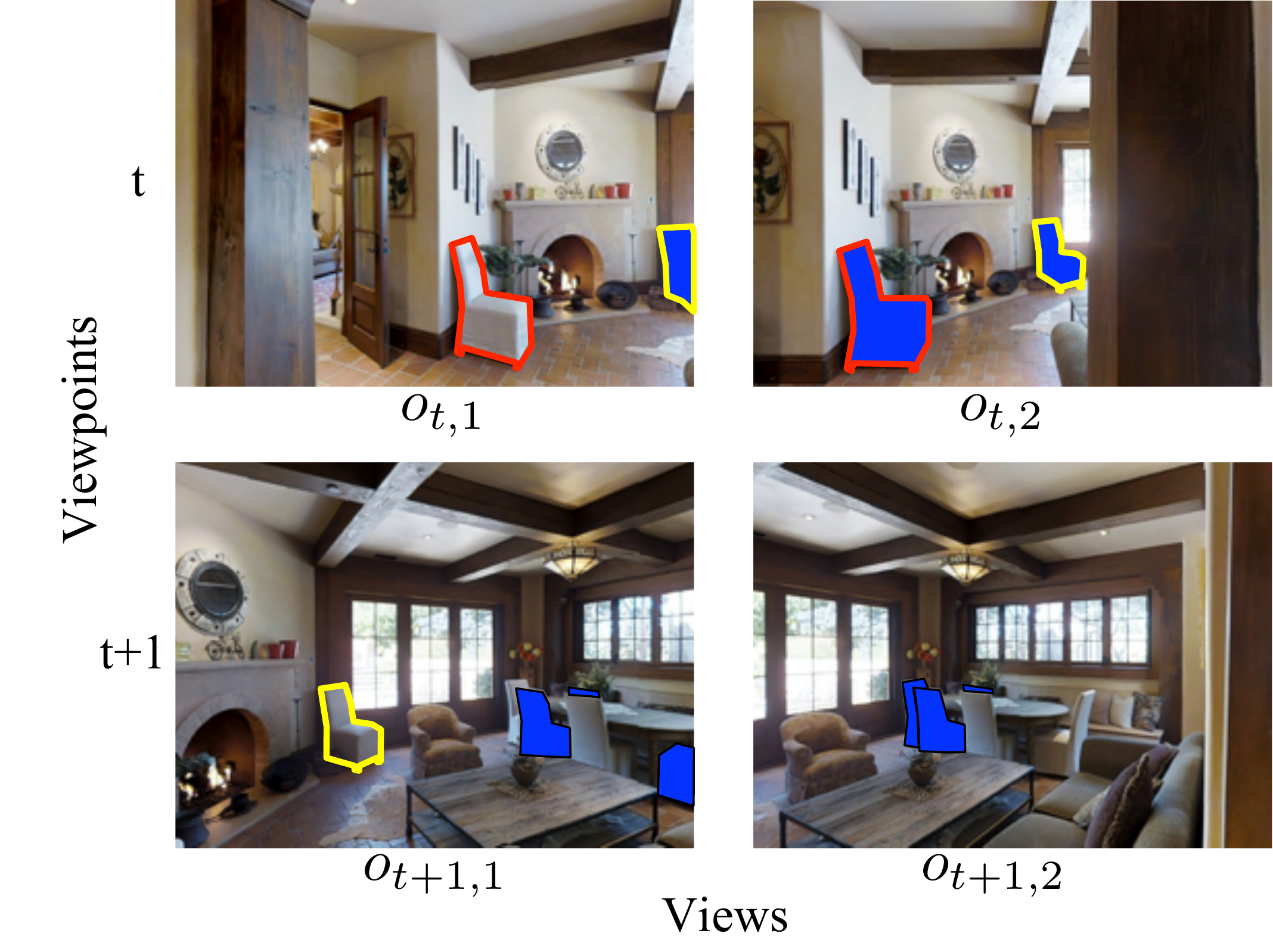}
        \subcaption{Feature dropout}
        \label{fig:feat_drop_illu}
    \end{subfigure} 
    ~
    \begin{subfigure}[t]{0.48\textwidth}
        \centering
        \includegraphics[width=1.0\textwidth]{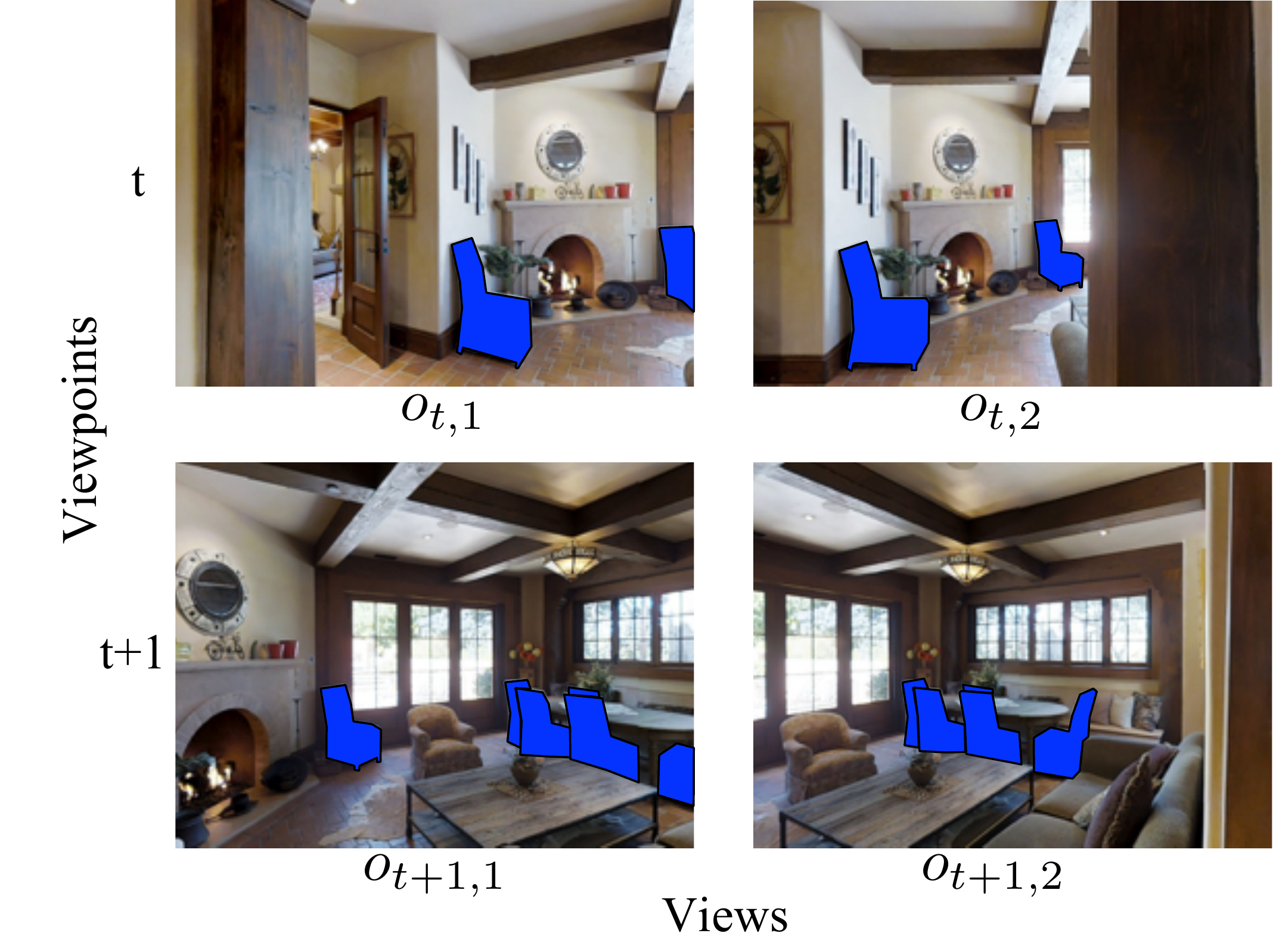}
        \subcaption{Environmental  dropout}
        \label{fig:env_drop_illu}
    \end{subfigure}
    \vspace{3pt}
    \caption{Comparison of the two dropout methods (based on an illustration on an RGB image).}
\end{figure*}

\begin{figure}[t]
    \centering
    \begin{subfigure}[t]{0.23\textwidth}
        \centering
        \includegraphics[width=0.99\textwidth]{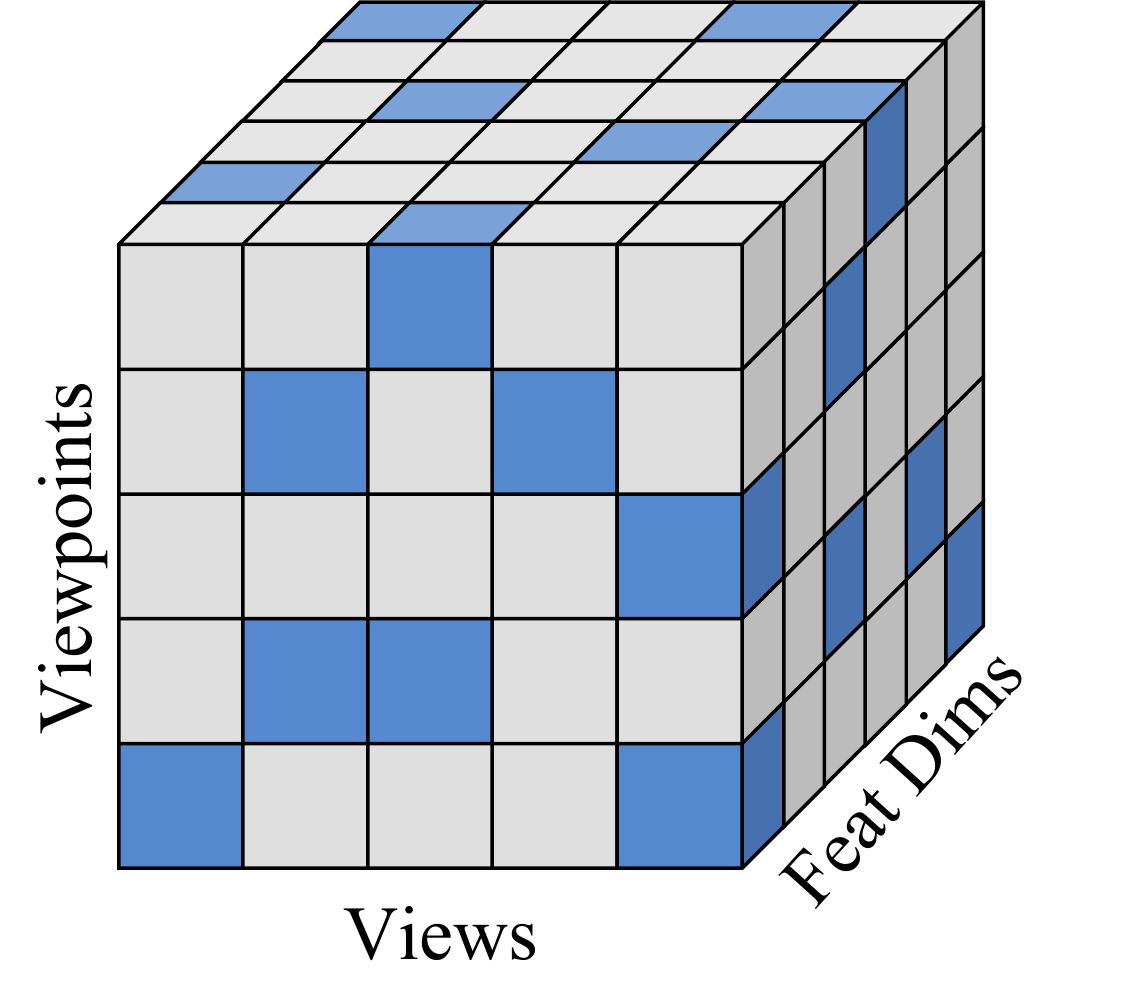}
        \subcaption{Feature dropout}
        \label{fig:feat_drop_cubic}
    \end{subfigure}%
    \begin{subfigure}[t]{0.23\textwidth}
        \centering
        \includegraphics[width=0.99\textwidth]{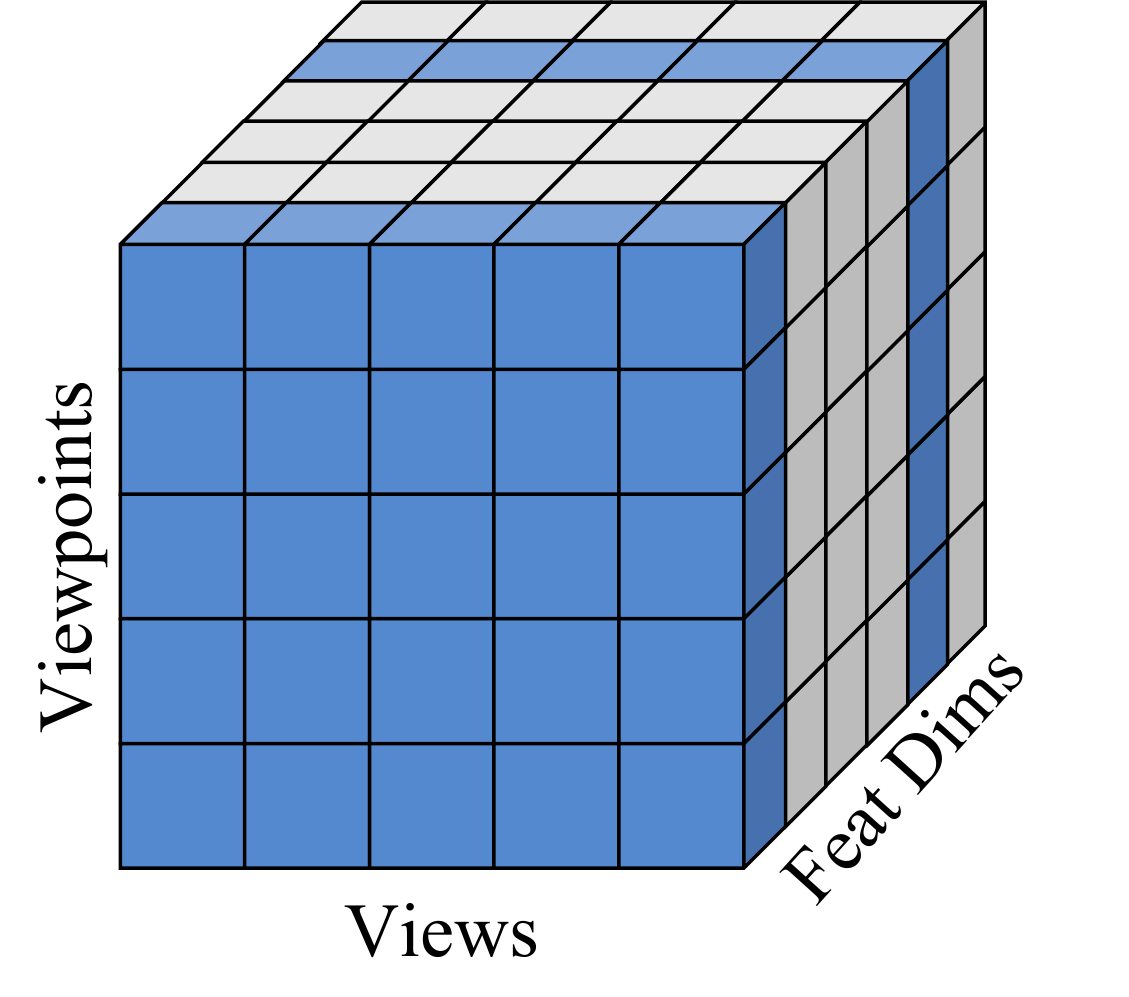}
        \subcaption{Environmental  dropout}
        \label{fig:env_drop_cubic}
    \end{subfigure}\\
    \vspace{4pt}
    \caption{Comparison of the two dropout methods (based on image features).}
    \vspace{-1pt}
\end{figure}

\subsection{Semi-Supervised Learning: Back Translation with Environmental Dropout}
\label{sec:ssl}

\subsubsection{Back Translation}  
\label{sec:back_translation}
Suppose the primary task is to learn the mapping of $\mathbf X \shortrightarrow \mathbf Y$ with paired data $\{(\mathbf X, \mathbf Y)\}$ and unpaired data $\{\mathbf{Y'}\}$.
In this case, the back translation method first trains a forward model $P_{\,\mathbf X \shortrightarrow \mathbf Y}$ and a backward model $P_{\,\mathbf Y \shortrightarrow \mathbf X}$, using paired data $\{(\mathbf X, \mathbf Y)\}$.
Next, it generates additional datum $\mathbf{X'}$ from the unpaired $\mathbf{Y'}$ using the backward model $P_{\,\mathbf Y \shortrightarrow \mathbf X}$.
Finally, $(\mathbf{X'}, \mathbf{Y'})$ are paired to further fine-tune the forward model $P_{\,\mathbf X \shortrightarrow \mathbf Y}$ as additional training data (also known as `data augmentation').

Back translation was introduced to the task of navigation in \newcite{fried2018speaker}. 
The forward model is a navigational agent $P_{\,\mathbf{E}, \mathbf{I} \shortrightarrow \mathbf R}$ (Sec.~\ref{sec:model}), which navigates inside an environment $\mathbf E$, trying to find the correct route $\mathbf R$ according to the given instruction $\mathbf I$.
The backward model is a \emph{speaker} $P_{\,\mathbf E, \mathbf R \shortrightarrow \mathbf{I}}$, 
which generates an instruction $\mathbf I$ from a given route $\mathbf R$ inside an environment $\mathbf{E}$. 
Our speaker model (details in Sec.~\ref{sec:speaker}) is an enhanced version of~\newcite{fried2018speaker}, where we use a stacked bidirectional LSTM-RNN encoder with attention flow.

For back translation, the Room-to-Room dataset labels around $7\%$ routes $\{ \mathbf R \}$ in the training environments\footnote{The number of all possible routes (shortest paths) in the $60$ existing training environments is 190K. Of these, the Room-to-Room dataset labeled around 14K routes with one navigable instruction for each, so the amount of labeled routes is around $7\%$ of 190K.}, so the rest of the routes $\{\mathbf{R'}\}$ are unlabeled.
Hence, we generate additional instructions $\mathbf{I'}$ using  $P_{\,\mathbf E, \mathbf R \shortrightarrow \mathbf{I}} \left(\mathbf E, \mathbf{R'} \right)$, so to obtain the new triplets $(\mathbf E, \mathbf{R'}, \mathbf{I'})$. 
The agent is then fine-tuned with this new data using the IL+RL method described in Sec.~\ref{sec:rl}. However, note that the environment $\mathbf E$ in the new triplet $(\mathbf E, \mathbf{R'}, \mathbf{I'})$ for semi-supervised learning is still selected from the \emph{seen} training environments. 
We demonstrate that the limited amount of environments $\{\mathbf E\}$ is actually the bottleneck of the agent performance in  Sec.~\ref{sec:necessary} and Sec.~\ref{sec:sufficient}.
Thus, we introduce our environmental dropout method to mimic the ``new'' environment $\mathbf{E'}$, as described next in Sec.~\ref{sec:env_drop}.

\subsubsection{Environmental Dropout} 
\label{sec:env_drop}
\paragraph{Failure of Feature Dropout}
Different from dropout on neurons to regularize neural networks,
we drop raw feature dimensions (see Fig.~\ref{fig:feat_drop_cubic}) to mimic the removal of random objects from an RGB image (see Fig.~\ref{fig:feat_drop_illu}).
This traditional feature dropout (with dropout rate $p$) is implemented as an element-wise multiplication of the feature $f$ and the dropout mask $\xi^f$.
Each element $\xi^f_e$ in the dropout mask $\xi^f$ is a sample of a random variable which obeys an independent and identical Bernoulli distribution multiplied by $1/(1-p)$.
And for different features, the distributions of dropout masks are independent as well.
\begin{align}
    \mathrm{dropout}_p(f) = & f \odot \xi^f \\
    \xi^f_e  \sim \frac{1}{1-p} & \mathrm{Ber}(1-p) 
\end{align}
Because of this independence among dropout masks, the traditional feature dropout fails in augmenting the existing environments because the `removal' is inconsistent in different views at the same viewpoint, and in different viewpoints. 

To illustrate this idea, we take the four RGB views in Fig.~\ref{fig:feat_drop_illu} as an example, where the chairs are randomly dropped from the views.
The removal of the left chair (marked with a red polygon) from view $o_{t,2}$ is inconsistent because it also appears in view $o_{t,1}$.
Thus, the speaker could still refer to it and the agent is aware of the existence of the chair.
Moreover, another chair (marked with a yellow polygon) is completely removed from viewpoint observation $o_t$, but the views in next viewpoint $o_{t+1}$ provides conflicting information which would confuse the speaker and the agent.
Therefore, in order to make generated environments consistent, we propose our environmental dropout method, described next. 
\paragraph{Environmental Dropout}
We create a new environment $\mathbf{E'}$ by applying environmental dropout on an existing environment $\mathbf{E}$.
\begin{equation}
    \mathbf {E'}   =  \mathrm{envdrop}_p(\mathbf{E})
\end{equation}
The view feature $f'_{t,i}$ observed from the new environment $\mathbf {E'}$ is calculated as an element-wise multiplication of the original feature $f_{t,i}$ and the environmental dropout mask $\xi^{\mathbf E}$ (see Fig.~\ref{fig:env_drop_cubic}):
\vspace{-5pt}
\begin{align}
    f'_{t,i}  & = f_{t,i} \odot \xi^{\mathbf E} \\
    \xi_e^{\mathbf E} & \sim \frac{1}{1-p} \mathrm{Ber}(1-p) \vspace{-5pt}
\end{align}
To maintain the spatial structure of viewpoints, only the image feature $\mathrm{ResNet}(v_{t,i})$ is dropped while the orientation feature $( \cos(\theta_{t,i}), \sin(\theta_{t,i}), \cos(\phi_{t,i}), \sin(\phi_{t,i}) )$ is fixed. 
As illustrated in Fig.~\ref{fig:env_drop_illu}, the idea behind environmental dropout is to mimic new environments by removing one specific class of object (e.g., the chair).
We demonstrate our idea by running environmental dropout on the ground-truth semantic views in Sec.~\ref{sec:feature_analysis}, which is proved to be far more effective than traditional feature dropout.
In practice, we perform the environmental dropout on image's visual feature where certain structures/parts are dropped instead of object instances, but the effect is similar.

We apply the environmental dropout to the back translation model as mentioned in Sec.~\ref{sec:back_translation}.
Note the environmental dropout method still preserves the connectivity of the viewpoints, thus we use the same way~\cite{fried2018speaker} to collect extra unlabeled routes $\{\mathbf{R'}\}$. 
We take \emph{speaker} to generate an additional instruction  $\mathbf{I'}\mbox{=} P_{\,\mathbf E, \mathbf R \shortrightarrow \mathbf{I}} \left(\mathbf {E'}, \mathbf{R'}\right)$ in the new environment $\mathbf{E'}$.
At last, we use IL+RL (in Sec.~\ref{sec:rl}) to fine-tune the model with this new triplet $(\mathbf{E'}, \mathbf{R'}, \mathbf{I'})$.

\subsubsection{Improvements on Speaker}
\label{sec:speaker}
Our speaker model is an enhanced version of the encoder-decoder model of~\newcite{fried2018speaker}, with improvements on the visual encoder: we stack two bi-directional LSTM encoders: a route encoder and a context encoder.
The route encoder takes features of ground truth actions $\{a^*_t\}_{t=1}^{T}$ from the route as inputs. 
Each hidden state $r_t$ then attends to surrounding views $\{f_{t,i}\}_{i=1}^{36}$ at each viewpoint.
The context encoder then reads the attended features and outputs final visual encoder representations:
\begin{align}
    r_1, ..., r_\textsc{t} & = \mathrm{Bi\mbox{-}LSTM}^\textsc{rte} (g_{1, a^*_1}, ..., g_{\textsc{t}, a^*_\textsc{t}}) \\
    \gamma_{t,i} &= \mathrm{softmax}_i(f_{t,i}^\intercal W_\textsc{r} \, r_{t}) \\
    \hat f_t &= \sum\nolimits_i \gamma_{t,i} f_{t,i} \\
    c_1, ..., c_\textsc{t} & = \mathrm{Bi\mbox{-}LSTM}^\textsc{ctx} (\hat f_1, ..., \hat f_\textsc{t})
\end{align}
The decoder is a regular attentive LSTM-RNN, which is discussed in Sec.~\ref{sec:model}. 
Empirically, our enhanced speaker model improves the BLEU-4 score by around 3 points.

\begin{table*}[t]
\small
\begin{center}
 \begin{tabular}{|l | c c c | c c c | c c c |}
  \hline
  Models  &  \multicolumn{9}{c|}{Test Unseen (Leader-Board)} \\
    & \multicolumn{3}{c|}{Single Run} & \multicolumn{3}{c|}{Beam Search} & \multicolumn{3}{c|}{Pre-Explore} \\

  \hline 
&  NL & SR(\%) & \emph{SPL} & NL & \emph{SR}(\%) & SPL &  NL & SR(\%) & \emph{SPL}   \\
\hline
Random~\cite{mattersim} & 	9.89 &	13.2 &	0.12 &	- &	- & - & - & - & - \\
Seq-to-Seq~\cite{mattersim} & 8.13 & 	20.4 &	0.18 &	- &	- & - & - & - & -\\
\hline
Look Before You Leap~\cite{wang2018look} &	9.15 &	25.3&	0.23&-  &-  &-  &-  &-  &- \\
Speaker-Follower~\cite{fried2018speaker} & 	14.8 &	35.0 &	0.28 & 1257 & 53.5  &\underline{0.01}  &-  &-  &- \\
Self-Monitoring~\cite{anonymous2019self-aware} & 18.0  &	\underline{48.0} &	0.35 &	373&    61.0 & \textbf{0.02}  &-  &- &-  \\
Reinforced Cross-Modal~\cite{wang2018reinforced} & 12.0 &43.1	&	\underline{0.38}&	358 & \underline{63.0} & \textbf{0.02}   &9.48&	60.5  &0.59  \\
\hline
Ours & 11.7  & \textbf{51.5}  & \textbf{0.47}  & 687 & \textbf{68.9}  & \underline{0.01} & 9.79 & \textbf{63.9}& \textbf{0.61}    \\
  \hline
\end{tabular}
\end{center}
\caption{
Leaderboard results under different experimental setups.
NL, SR, and SPL are Navigation Length, Success Rate and Success rate weighted by Path Length.
The primary metric for each setup is in italics. 
The best results are in bold font and the second best results are underlined.
}
\vspace{-9pt}
\label{table:result}
\end{table*}
\section{Experimental Setup}

\paragraph{Dataset and Simulator}
\label{sec:data}
We evaluate our agent on the Matterport3D simulator \cite{mattersim}. 
Navigation instructions in the dataset are collected via Amazon Mechanical Turk by showing them the routes in the Matterport3D environment~\cite{Matterport3D}. 
The dataset is split into training set (61 environments, 14,025 instructions), seen validation set (61 environments, 1,020 instructions), unseen validation set (11 environments, 2,349 instructions), and unseen test set (18 environments, 4,173 instructions). 
The unseen sets only involve the environments outside the training set. 

\paragraph{Evaluation Metrics}
For evaluating our model, Success Rate (SR) is the primary metric. 
The execution route by the agent is considered a success when the navigation error is less than $3$ meters. 
Besides success rate, we use three other metrics\footnote{
The Oracle Success Rate (OSR) is not included because it's highly correlated with the Navigation Length.
}
: Navigation Length (NL), Navigation Error (NE),  and Success rate weighted by Path Length (SPL) \cite{anderson2018evaluation}.
Navigation Error (NE) is the distance between target viewpoint $\mathbf{T}$ and agent stopping position.

\paragraph{Implementation Details}
\label{sec:train_detail}
Similar to the traditional dropout method, the environmental dropout mask is computed and applied at each training iteration.
Thus, the amount of unlabeled semi-supervised data used is not higher in our dropout method.
We also find that sharing the environmental dropout mask in different environments inside a batch will stabilize the training.
To avoid over-fitting, the model is early-stopped according to the success rate on the unseen validation set. More training details in appendices.
\section{Results}
\label{sec:result}
In this section, we compare our agent model with the models in previous works on the Vision and Language Navigation (VLN) leaderboard.
The models on the leaderboard are evaluated on a private unseen test set which contains $18$ new environments.
We created three columns in Table~\ref{table:result} for different experimental setups: single run, beam search, and unseen environments pre-exploration. 
For the result, our model outperforms all other models in all experimental setups.

\paragraph{Single Run}
Among all three experimental setups, single run is the most general and highly correlated to the agent performance. Thus, it is considered as the primary experimental setup.
In this setup, the agent navigates the environment once and is not allowed\footnote{According to the Vision and Language Navigation (VLN) challenge submission guidelines} to:
(1) run multiple trials,
(2) explore nor map the test environments before starting.
Our result is $3.5\%$ and $9\%$ higher than the second-best in Success Rate and SPL, resp. 

\paragraph{Beam Search}
In the beam search experimental setup, 
an agent navigates the environment, collects multiple routes, re-ranks them, and selects the route with the highest score as the prediction.
Besides showing an upper bound,
beam search is usable when the environment is explored and saved in the agent's memory but the agent does not have enough computational capacity to fine-tune its navigational model. 
We use the same beam-search algorithm, state factored Dijkstra algorithm, to navigate the unseen test environment. 
Success Rate of our model is $5.9\%$ higher than the second best.
SPL metric generally fails in evaluating beam-search models because of the long Navigation Length (range of SPL is $0.01$-$0.02$).

\paragraph{Pre-Exploration}
The agent pre-explores the test environment before navigating and updates its agent model with the extra information.
When executing the instruction in the environment, the experimental setup is still ``single run''.
The ``pre-exploration'' agent mimics the domestic robots (e.g., robot vacuum) which only needs to navigate the seen environment most of the time.
For submitting to the leaderboard, we simply train our agent via back translation with environmental dropout on test unseen environments (see Sec.\ref{sec:sufficient}).
Our result is $3.4\%$ higher than \newcite{wang2018reinforced} in Success Rate and $2.0\%$ higher in SPL.
\footnote{To fairly compare with \newcite{wang2018reinforced}, we exclude the exploration route in calculating Navigation Length.}

\begin{table*}[t]
\small
\begin{center}

 \begin{tabular}{|l | c c c c | c c c c|}
  \hline
  Models & \multicolumn{4}{c|}{Val Seen} &  \multicolumn{4}{c|}{Val Unseen} \\
  \hline 
   & NL(m) & NE(m)  & SR(\%) & SPL &   NL(m) & NE(m)  & SR(\%) & SPL  \\
\hline\hline
\multicolumn{9}{|c|}{SUPERVISED LEARNING}\\
\hline\hline
Behavioral Cloning (IL) & 10.3&	5.39&	48.4&	0.46&	9.15&	6.25&	43.6&	0.40\\
Advantage Actor-Critic (RL) & 73.8&	7.11&	22.0&	0.03&	73.8&	7.32&	24.0&	0.03\\
IL + RL& 10.1&	4.71&	55.3&	0.53&	9.37&	5.49&	46.5&	0.43 \\
\hline\hline
\multicolumn{9}{|c|}{ SEMI-SUPERVISED LEARNING }\\
\hline\hline
Back Translation & 10.3&	4.19&	58.1&	0.55&	10.5&	5.43&	48.2&	0.44 \\
+ Feat Drop & 10.3&	4.13&	58.4&	0.56& 	9.62&	5.43&	48.4&	0.45\\
+ Env Drop (No Tying)  &10.3&	4.32&	57.3&	0.55&	9.51&	5.27&	49.0&	0.46 \\
+ Env Drop (Tying)&  11.0&	3.99&	62.1&	0.59&	10.7&	5.22&	52.2&	0.48 \\ 
\hline\hline
\multicolumn{9}{|c|}{ FULL MODEL }\\
  \hline\hline
 Single Run&  11.0&	3.99&	62.1&	0.59&	10.7&	5.22&	52.2&	0.48\\
  Beam Search  &703 &	2.52 &	75.7 &	0.01&	663&	3.08&	69.0&	0.01 \\
  Pre-Explore  & 9.92 &	4.84 & 54.7 &	0.52&	9.57&	3.78&	64.5&	0.61 \\
  \hline
\end{tabular}
\end{center}
\caption{For the ablation study, we show the results of our different methods on validation sets. 
Our full model (single run) gets $8.6\%$ improvement in validation unseen success rate above our baseline. 
And both the supervised learning (IL+RL) and semi-supervised learning methods (back translation + env drop) have substantial contributions to our final result.
}
\vspace{-5pt}
\label{table:ablation}
\end{table*}

\section{Ablation Studies}
\label{sec:ablations}

\paragraph{Supervised Learning}
We first show the effectiveness of our IL+RL method by comparing it with the baselines (Table~\ref{table:ablation}).
We implement Behavioural Cloning\footnote{
The Behavioral Cloning (IL) baseline is the same as the panoramic view baseline in \newcite{fried2018speaker} except for two differences:
(1) The agent takes the teacher action instead of the sampled action from the distribution (see ``imitation learning'' of Sec.~\ref{sec:rl}),
(2) The hidden input of the LSTM is the instruction-aware hidden from the previous step (see Sec.~\ref{sec:model}).
We improve our baseline result with these modifications.}
and Advantage Actor-Critic as our imitation learning (IL) and reinforcement learning (RL) baselines, respectively.
The mixture of IL+RL (see Sec.~\ref{sec:rl}) outperforms the IL-only model and RL-only model by $2.9\%$ and $22.5\%$, 
which means that our IL+RL could overcome the misleading teacher actions in IL and significantly stabilize the training of RL.

\paragraph{Semi-Supervised Learning}
We then fine-tune our best supervised model (i.e., IL+RL) with back translation.
Besides providing a warm-up, IL+RL is also used to learn the new generated data triplets in back translation.
As shown in Table~\ref{table:ablation}, back translation with environmental dropout improves the best supervised model by $5.7\%$, where the improvement is $3$ times more than the back translation without new environments.
We then show the results of the alternatives to environmental dropout.
The performance with feature dropout is almost the same to the original back translation, which is $3.8\%$ lower than the environmental dropout. 
We also prove that the improvement from the environmental dropout method does not only come from generating \emph{diverse} instructions introduced by dropout in the speaker, but also comes from using the same dropout mask in the follower agent too.
To show this, we use two independent (different) environmental dropout masks for the speaker and the follower (i.e., no tying of the dropout mask), and the result drops a lot as compared to when we tie the speaker and follower dropout masks.

\paragraph{Full Model}
Finally, we show the performance of our best agent under different experimental setups.
The ``single run'' result is copied from the best semi-supervised model for comparison. 
The state-factored Dijkstra algorithm \cite{fried2018speaker} is used for the beam search result.
The method for pre-exploration is described in Sec.~\ref{sec:sufficient}, where the agent applies back translation with environmental dropout on the validation unseen environment.

\section{Analysis}
\label{sec:analysis}
In this section, we present analysis experiments that first exposed the limited environments bottleneck to us, and hence inspired us to develop our environmental dropout method to break this bottleneck.
\begin{figure}[t]
  \includegraphics[width=0.5\textwidth]{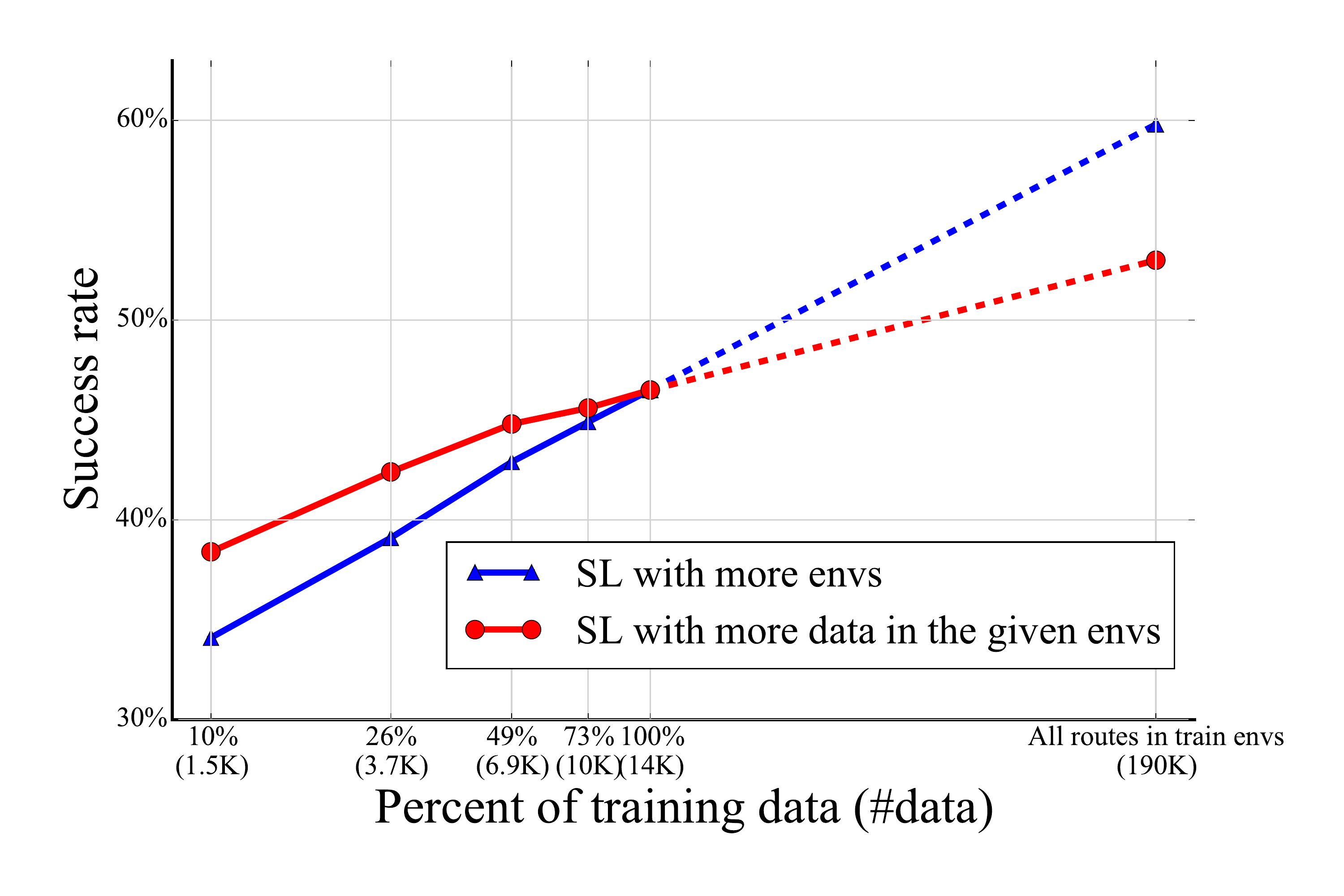} 
\caption{
Success rates of agents trained with different amounts of data. X-axis in log-scale.
The blue line represents the growth of results by gradually adding new environments to the supervised training method.
The red line is trained with the same amounts of data as the blue line, but the data is randomly selected from all $60$ training environments.
The dashed lines are predicted.
}
\label{fig:analysis}
\vspace{-10pt}
\end{figure}

\subsection{More Environments vs. More Data}
\label{sec:necessary}
In order to show that more environments are crucial for  better performance of agents, in Fig.~\ref{fig:analysis}, we present the result of Supervised Learning (SL) with different amounts of data selected by two different data-selection methods.
The first method gradually uses more \#environments (see the blue line ``SL with more envs'') while the second method selects data from the whole training data with all 60 training environments (see the red line ``SL with more data''). Note that the amounts of data in the two setups are the same for each plot point
As shown in Fig.~\ref{fig:analysis}, the ``more envs'' selection method shows \emph{higher growth rate} in success rate than the ``more data'' method. 
We also predict the success rates (in dashed line) with the prediction method in \newcite{sun2017revisiting}.
The predicted result is much higher when training with more environments.
The predicted result (the right end of the red line) also shows that the upper bound of Success Rate is around $52\%$ if all the 190K routes in the training environments is labeled by human (instead of being generated by \textit{speaker} via back translation), which indicates the need for ``new'' environments.

\subsection{Back Translation on Unseen Environments}
\label{sec:sufficient}
In this subsection, we show that back translation could significantly improve the performance when it uses new data triplets from testing environments --- the unseen validation environments where the agent is evaluated in.
Back translation (w.o. Env Drop) on these unseen environments achieves a success rate of $61.9\%$, while the back translation on the training environments only achieves $46.5\%$.
The large margin between the two results indicates the need of ``new'' environments in back translation.
Moreover, our environmental dropout on testing environments could further improve the result to $64.5\%$, which means that the amount of environments in back translation is far from enough.

\begin{figure}[t!]
    \centering
    \begin{subfigure}[t]{0.24\textwidth}
        \centering
        \includegraphics[width=0.95\textwidth]{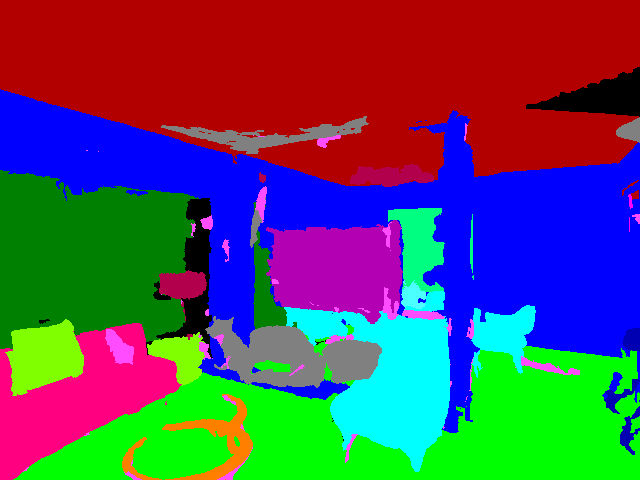}
        \subcaption{Semantic View}
    \end{subfigure}%
    \begin{subfigure}[t]{0.24\textwidth}
        \centering
        \includegraphics[width=0.95\textwidth]{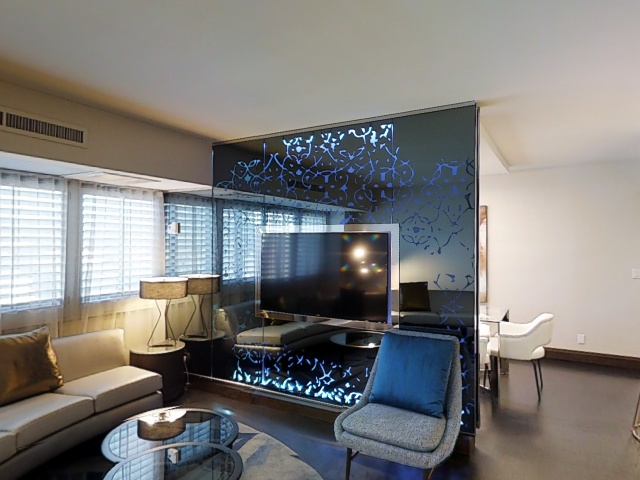}
        \subcaption{RGB view}
    \end{subfigure}
    \vspace{1pt}
    \caption{Comparison of semantic and raw RGB views.
}
    \label{fig:sem_view}
    \vspace{-6pt}
\end{figure}
\subsection{Semantic Views}
\label{sec:feature_analysis}
To demonstrate our intuition of the success of environmental dropout (in Sec.~\ref{sec:env_drop}), we replace the image feature $\mathrm{ResNet}(v_{t,i})$ with the semantic view feature.
The semantic views (as shown in Fig.~\ref{fig:sem_view}) are rendered from the Matterport3D dataset \cite{Matterport3D}, where different colors indicate different types of objects.
Thus, dropout on the semantic view feature would remove the object from the view.
With the help of this additional information (i.e., the semantic view), the success rate of IL+RL is $49.5\%$ on the unseen validation set.
Back translation (without dropout) slightly improves the result to $50.5\%$.
The result with feature dropout is $50.2\%$ while the environmental dropout could boost the result to $52.0\%$, which supports our claim in Sec.~\ref{sec:env_drop}.

\section{Conclusion}
We presented a navigational agent which better generalizes to unseen environments.
The agent is supervised with a mixture of imitation learning and reinforcement learning.
Next, it is fine-tuned with semi-supervised learning, with speaker-generated instructions. Here, we showed that the limited variety of environments is the bottleneck of back translation and we overcome it via `environmental dropout' to generate new unseen environments.
We evaluate our model on the Room-to-Room dataset and  achieve rank-1 in the Vision and Language Navigation (VLN) challenge leaderboard under all experimental setups.

\section*{Acknowledgments}
We thank the reviewers for their helpful comments. 
This work was supported by ARO-YIP Award \#W911NF-18-1-0336, ONR Grant \#N00014-18-1-2871, and faculty awards from Google, Facebook, Adobe, Baidu, and Salesforce. The views, opinions, and/or findings contained in this article are those of the authors and should not be interpreted as representing the official views or policies, either expressed or implied, of the funding agency.

\bibliography{naaclhlt2019}
\bibliographystyle{acl_natbib}

\appendix

\section{Appendices}
\subsection{Implementation Details}
We use ResNet-152 \cite{he2016deep} pretrained on the ImageNet \cite{russakovsky2015imagenet} to extract the $2048$-dimensional image feature. 
The agent model is first trained with supervised learning via the mixture of imitation and reinforcement learning. 
The model is then fine-tuned by back translation with environmental dropout. 
To stabilize the optimization of back translation, we calculate supervised loss for half of the batch and semi-supervised loss for the other half.
We find that sharing the environmental dropout mask in different environments inside the same batch will stabilize the training.

The word embedding is trained from scratch with size $256$ and the dimension of the action embedding is $64$. 
The size of the LSTM units is set to $512$ ($256$ for the bidirectional LSTM).
In RL training, the discounted factor $\gamma$ is $0.9$. 
We use the reward shaping \cite{wu2018building}: the direct reward $r_t$ at time step $t$ is the change of the distance $d_{t-1} - d_t$, supposing $d_t$ is the distance to the target position at time step $t$. 
The maximum decoding action length is set to $35$.
For optimizing the loss, we use RMSprop \cite{hinton2012neural} (as suggested in~\newcite{mnih2016asynchronous}) with a fixed learning rate $1e-4$ and the batch size is $64$. 
We applying dropout rate $0.4$ to the environmental dropout and $0.5$ to the dropout layers which regularize the network.
The global gradient norm is clipped by $40$. 
We tuned the hyper-parameters based on the Success Rate of the unseen validation set.

When working with the semantic view, the \emph{key} labels (e.g., \emph{wall}, \emph{floor}, \emph{ceil}) are not dropped, because they are the basic structure of the environment. Empirically, no improvement will be achieved when the \emph{key} labels are dropped as well.

\end{document}